\documentclass[conference]{IEEEtran}  

\usepackage[english]{babel}
\usepackage[utf8x]{inputenc}
\usepackage[T1]{fontenc}

\usepackage{amsmath}
\usepackage{bbm}
\usepackage{graphicx}
\usepackage[colorinlistoftodos]{todonotes}
\usepackage{hyperref}

\usepackage{url}            
\usepackage{amsfonts}       
\usepackage{nicefrac}       
\usepackage{mathtools}
\usepackage{amsmath}
\usepackage{algorithm}
\usepackage{algorithmic}
\usepackage{booktabs}

\newcommand\blfootnote[1]{%
  \begingroup
  \renewcommand\thefootnote{}\footnote{#1}%
  \addtocounter{footnote}{-1}%
  \endgroup
}

\usepackage{graphicx}
\usepackage[font={small}]{caption}
\usepackage{subcaption}

\newcommand{\cM}{\mathcal{M}}

\newcommand{\cS}{\mathcal{S}}
\newcommand{\cA}{\mathcal{A}}
\newcommand{\cT}{\mathcal{T}}
\newcommand{\cR}{\mathcal{R}}

\newcommand{\bR}{\mathbb{R}}
\newcommand{\bE}{\mathbb{E}}

\newcommand{\Vpi}{V^{\pi}}
\newcommand{\Qpi}{Q^{\pi}}
\newcommand{\Api}{A^{\pi}}
\newcommand{\pith}{\pi_\theta}

\usepackage{authblk}

\title{Learning Complex Dexterous Manipulation with Deep Reinforcement Learning and Demonstrations}
\author{Aravind Rajeswaran${^{1*}}$\thanks{* These authors contributed equally to this work.}, Vikash Kumar${^{1,2*}}$, Abhishek Gupta${^3}$, Giulia Vezzani${^4}$, \\ John Schulman${^2}$, Emanuel Todorov${^1}$, Sergey Levine${^3}$}

\begin{document}
\maketitle 
\blfootnote{$\;^*$ Equal contributions. $^1$~University of Washington Seattle, $^2$~OpenAI, $^3$~University of California Berkeley, $^4$~Istituto Italiano di Tecnologia. Correspond to $\lbrace$ aravraj, vikash $\rbrace$ @ cs.washington.edu}

\begin{abstract}
Dexterous multi-fingered hands are extremely versatile and provide a generic way to perform a multitude of tasks in human-centric environments. However, effectively controlling them remains challenging due to their high dimensionality and large number of potential contacts. Deep reinforcement learning (DRL) provides a model-agnostic approach to control complex dynamical systems, but has not been shown to scale to high-dimensional dexterous manipulation. Furthermore, deployment of DRL on physical systems remains challenging due to sample inefficiency. Consequently, the success of DRL in robotics has thus far been limited to simpler manipulators and tasks. In this work, we show that model-free DRL can effectively scale up to complex manipulation tasks with a high-dimensional 24-DoF hand, and solve them from scratch in simulated experiments. Furthermore, with the use of a small number of human demonstrations, the sample complexity can be significantly reduced, which enables learning with sample sizes equivalent to a few hours of robot experience. The use of demonstrations result in policies that exhibit very natural movements and, surprisingly, are also substantially more robust. We demonstrate successful policies for object relocation, in-hand manipulation, tool use, and door opening, which are shown in the supplementary video.
\end{abstract}
\IEEEpeerreviewmaketitle

\section{Introduction}

Multi-fingered dexterous manipulators are crucial for robots to function in human-centric environments, due to their versatility and potential to enable a large variety of contact-rich tasks, such as in-hand manipulation, complex grasping, and tool use. However, this versatility comes at the price of high dimensional observation and action spaces, complex and discontinuous contact patterns, and under-actuation during non-prehensile manipulation. This makes dexterous manipulation with multi-fingered hands a challenging problem.

Dexterous manipulation behaviors with multi-fingered hands have previously been obtained using model-based trajectory optimization methods~\cite{mordatch2012contact, kumar2014real}. However, these methods typically rely on accurate dynamics models and state estimates, which are often difficult to obtain for contact rich manipulation tasks, especially in the real world. Reinforcement learning provides a model agnostic approach that circumvents these issues. Indeed, model-free methods have been used for acquiring manipulation skills~\cite{vanHoof2015learning,shaneNAF}, but so far have been limited to simpler behaviors with 2-3 finger hands or whole-arm manipulators, which do not capture the challenges of high-dimensional multi-fingered hands.

\begin{figure}[!t]
\centering
\includegraphics[width=0.45\columnwidth]{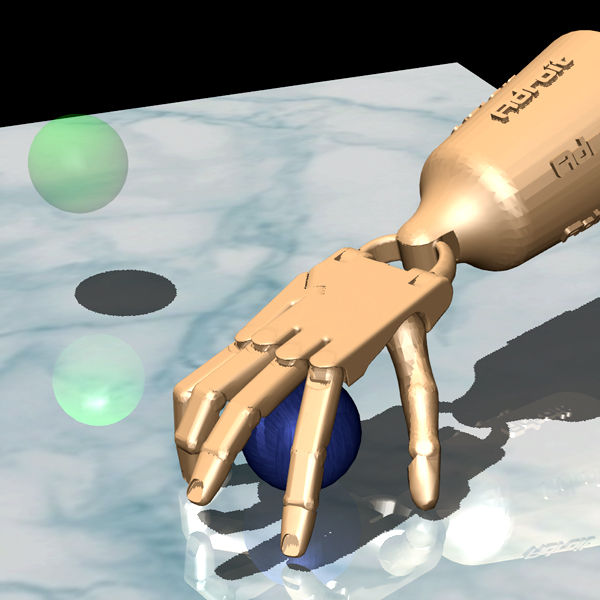}
\includegraphics[width=0.45\columnwidth]{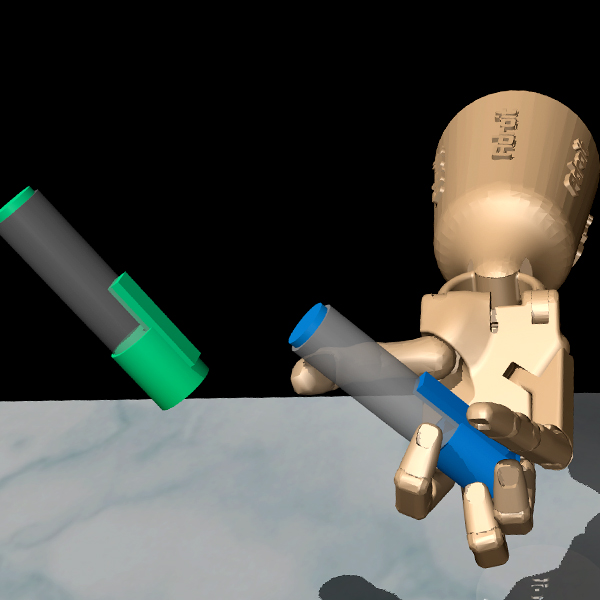}\\
 \vspace{.1cm}
\includegraphics[width=0.45\columnwidth]{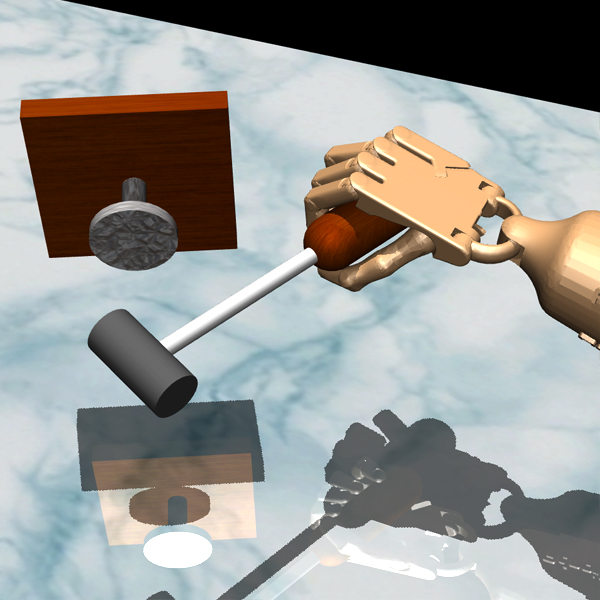}
\includegraphics[width=0.45\columnwidth]{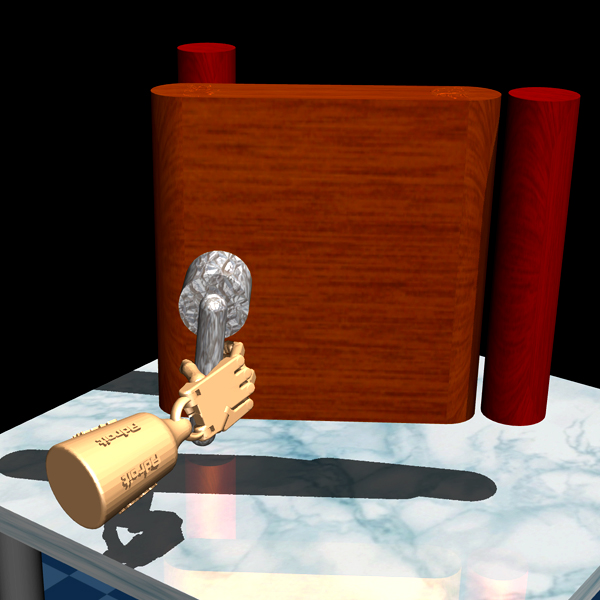}
\caption{We demonstrate a wide range of dexterous manipulation skills such as object relocation, in-hand manipulation, tool use, and opening doors using DRL methods. By augmenting with human demonstrations, policies can be trained in the equivalent of a few real-world robot hours.}
\vspace{-0.5cm}
        \label{fig:behaviors}
\end{figure}

DRL research has made significant progress in improving performance on standardized benchmark tasks, such as the OpenAI gym benchmarks~\cite{gym}. However, the current benchmarks are typically quite limited both in the dimensionality of the tasks and the complexity of the interactions. Indeed, recent work has shown that simple linear policies are capable of solving many of the widely studied benchmark tasks~\cite{Rajeswaran17}. Thus, before we can develop DRL methods suitable for dexterous manipulation with robotic hands, we must set up a suite of manipulation tasks that exercise the properties that are most crucial for real-world hands: high dimensionality, rich interactions with objects and tools, and sufficient task variety. To that end, we begin by proposing a set of 4 dexterous manipulation tasks in simulation, which are illustrated in Figure~1. These tasks are representative of the type of tasks we expect robots to be proficient at: grasping and moving objects, in-hand manipulation, and tool usage among others. Using these representative tasks, we study how DRL can enable learning of dexterous manipulation skills. Models and code accompanying this work can be found at: \url{http://sites.google.com/view/deeprl-dexterous-manipulation}

We find that existing RL algorithms can indeed solve these dexterous manipulation tasks, but require significant manual effort in reward shaping. In addition, the sample complexity of these methods is very poor, thus making real world training infeasible, and the resulting policies exhibit idiosyncratic strategies and poor robustness. To overcome this challenge, we propose to augment the policy search process with a small number of human demonstrations collected in virtual reality (VR). In particular, we find that pre-training a policy with behavior cloning, and subsequent fine-tuning with policy gradient along with an augmented loss to stay close to the demonstrations, dramatically reduces the sample complexity, enabling training within the equivalent of a few real-world robot hours. The use of human demonstrations also provides additional desirable qualities such as human-like smooth behaviors and robustness to variations in the environment. Although success remains to be demonstrated on hardware, our results in this work indicate that DRL methods when augmented with demonstrations are a viable option for real-world learning of dexterous manipulation skills. Our contributions are:

\begin{itemize}
\item We demonstrate, in simulation, dexterous manipulation with high-dimensional human-like five-finger hands using model-free DRL. To our knowledge, this is the first empirical result that demonstrates model-free learning of tasks of this complexity.
\item We show that with a small number of human demonstrations, the sample complexity can be reduced dramatically and brought to levels which can be executed on physical systems.
\item We also find that policies trained with demonstrations are more human-like as well as robust to variations in the environment. We attribute this to human priors in the demonstrations which bias the learning towards more robust strategies.
\item We propose a set of dexterous hand manipulation tasks, which would be of interest to researchers at the intersection of robotic manipulation and machine learning. 
\end{itemize}

\section{Related Work} 

Manipulation with dexterous hands represents one of the most complex and challenging motor control tasks carried out by humans, and replicating this behavior on robotic systems has proven extremely difficult. Aside from the technical challenges of constructing multi-fingered dexterous hands, control of these manipulators has turned out to be exceedingly challenging. Many of the past successes in dexterous manipulation have focused either on designing hands that mechanically simplify the control problem~\cite{RBOorginal,softHand}, at the expense of reduced flexibility, or on acquiring controllers for relatively simple behaviors such as grasping~\cite{petersgrasping} or rolling objects in the fingers~\cite{vanHoof2015learning}, often with low degree of freedom manipulators. Our work explores how a combination of DRL and demonstration can enable dexterous manipulation with high-dimensional human-like five-finger hands~\cite{kumar2013fast,xu2013low}, controlled at the level of individual joints. We do not aim to simplify the morphology, and explore highly complex tasks that involve tool use and object manipulation. Although our experiments are in simulation, they suggest that DRL might in the future serve as a powerful tool to enable much more complex dexterous manipulation skills with complex hands.

\paragraph{\bf Model-based trajectory optimization}:
Model-based trajectory optimization methods~\cite{mordatch2012contact,kumar2014real,PosaCT14} have demonstrated impressive results in simulated domains, particularly when the dynamics can be adjusted or relaxed to make them more tractable (as, e.g., in computer animation). Unfortunately, such approaches struggle to translate to real-world manipulation since prespecifying or learning complex models on real world systems with significant contact dynamics is very difficult. Although our evaluation is also in simulation, our algorithms do not make any assumption about the structure of the dynamics model, requiring only the ability to generate sample trajectories. Our approach can be executed with minimal modification on real hardware, with the limitation primarily being the number of real-world samples required. As we will show, this can be reduced significantly with a small number of demonstrations, suggesting the possibility of performing learning directly in the real world. 

\paragraph{\bf Model-free reinforcement learning}
Model-free RL methods~\cite{sutton1998reinforcement,Theodorou2010AGP} and versions with deep function approximators~\cite{Schulman15,ddpg} do not require a model of the dynamics, and instead optimize the policy directly. However, their primary drawback is the requirement for a large number of real-world samples. Some methods like PoWER overcome this limitation through the use of simple policy representations such as DMPs and demonstrate impressive results~\cite{DMP2008, KoberP11}. However, more complex representations may be needed in general for more complex tasks and to incorporate rich sensory information. More recently, RL methods with rich neural network function approximators have been studied in the context of basic manipulation tasks with 7-10 DoF manipulators, and have proposed a variety of ways to deal with the sample complexity. 
Prior work has demonstrated model-free RL in the real world with simulated pre-training~\cite{KTHpikachu} and parallelized data collection~\cite{shaneNAF} for lower-dimensional whole-arm manipulation tasks.
Our work builds towards model-free DRL on anthropomorphic hands, by showing that we can reduce sample complexity of learning to practical levels with a small number of human demonstrations.
Prior work has demonstrated learning of simpler manipulation tasks like twirling a cylinder with a similar morphology~\cite{kumar2016optimal} using guided policy search~\cite{Levine16}, and extended this approach to also incorporate demonstrations~\cite{kumar2016experience}.
However, the specific tasks we demonstrate are substantially more complex featuring a large number of contact points and tool use, as detailed in Section~\ref{sec:tasks}.

\paragraph{\bf Imitation learning}
In imitation learning, demonstrations of successful behavior are used to train policies that imitate the expert providing these successful trajectories~\cite{schaalimitation}. A simple approach to imitation learning is behavior cloning (BC), which learns a policy through supervised learning to mimic the demonstrations. Although BC has been applied successfully in some instances like autonomous driving~\cite{nvidiaimitation}, it suffers from problems related to distribution drift~\cite{dagger}. 
Furthermore, pure imitation learning methods cannot exceed the capabilities of the demonstrator since they lack a notion of task performance. In this work, we do not just perform imitation learning, but instead use imitation learning to bootstrap the process of reinforcement learning. The bootstrapping helps to overcome exploration challenges, while RL fine-tuning allows the policy to improve based on actual task objective. 

\paragraph{\bf Combining RL with demonstrations}
Methods based on dynamic movement primitives (DMPs)~\cite{DMP2008,PIDMP,DMPLWR, Ijspeert} have been used to effectively combine demonstrations and RL to enable faster learning. Several of these methods use trajectory-centric policy representations, which although well suited for imitation, do not enable feedback on rich sensory inputs. Although such methods have been applied to some dexterous manipulation tasks~\cite{vanHoof2015learning}, the tasks are comparatively simpler than those illustrated in our work. Using expressive function approximators allow for complex, nonlinear ways to use sensory feedback, making them well-suited to dexterous manipulation.

In recent work, demonstrations have been used for pre-training a Q-function by minimizing TD error~\cite{DQfD}. Additionally demonstrations have been used to guide exploration through reward/policy shaping but these are often rule-based or work on discrete spaces making them difficult to apply to high dimensional dexterous manipulation~\cite{HAT,brys,sub}. The work most closely related to ours is DDPGfD~\cite{DDPGfD}, where demonstrations are incorporated into DDPG~\cite{ddpg} by adding them to the replay buffer. This presents a natural and elegant way to combine demonstrations with an off-policy RL method. In concurrent work~\cite{ashvin}, this approach was combined with hindsight experience replay~\cite{her}. The method we propose in this work bootstraps the policy using behavior cloning, and combines demonstrations with an on-policy policy gradient method. Off-policy methods, when successful, tend to be more sample efficient, but are generally more unstable~\cite{rllab,deeprlthatmatters}. On-policy methods on the other hands are more stable, and scale well to high dimensional spaces~\cite{rllab}.
Our experimental results indicate that with the incorporation of demonstrations, the sample complexity of on-policy methods can be dramatically reduced, while retaining their stability and robustness. Indeed, the method we propose significantly outperforms DDPGfD. 
Concurrent works with this paper have also proposed to integrate demonstrations into reward functions~\cite{Peng2018}, and there have been attempts to learn with imperfect demonstrations~\cite{Yang2018}.
Overall, we note that the general idea of bootstrapping RL with supervised training is not new. However, the extent to which it helps with learning of complex dexterous manipulation skills is surprising and far from obvious.

\section{Dexterous Manipulation Tasks}
\label{sec:tasks}
The real world presents a plethora of interesting and important manipulation tasks. While solving individual tasks via custom manipulators in a controlled setting has led to success in industrial automation, this is less feasible in an unstructured settings like the home. Our goal is to pick a minimal task-set that captures the technical challenges representative of the real world. We present four classes of tasks - object relocation, in-hand manipulation, tool use, and manipulating environmental props (such as doors). Each class exhibits distinctive technical challenges, and represent a large fraction of tasks required for proliferation of robot assistance in daily activities -- thus being potentially interesting to researchers at the intersection of robotics and machine learning. All our task environments expose hand (joint angles), object (position and orientation), and target (position and orientation) details as observations, expect desired position of hand joints as actions, and provides an oracle to evaluate success. We now describe the four classes in light of the technical challenges they present. 

\begin{figure}[t!]
  \centering
  \includegraphics[width=0.2375\columnwidth]{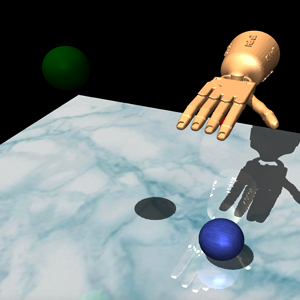} 
  \includegraphics[width=0.2375\columnwidth]{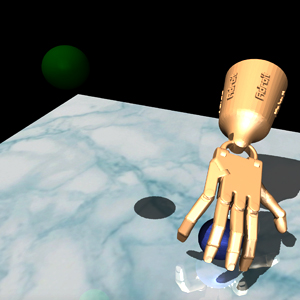} 
  \includegraphics[width=0.2375\columnwidth]{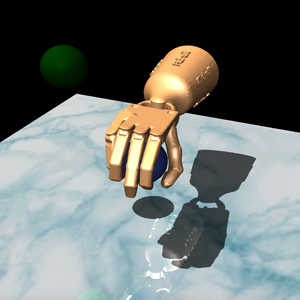} 
  \includegraphics[width=0.2375\columnwidth]{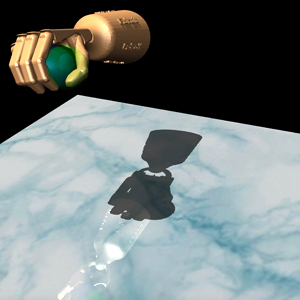} 
  \footnotesize\caption{Object relocation -- move the blue ball to the green target. Positions of the ball and target are randomized over the entire workspace. Task is considered successful when the object is within epsilon-ball of the target.}
  \label{fig:pickup}
\end{figure}

\begin{figure}[t!]
  \centering
  \includegraphics[width=0.2375\columnwidth]{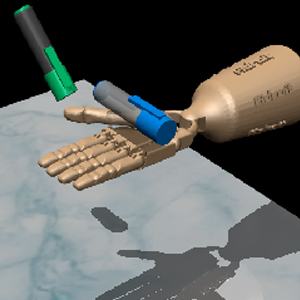} 
  \includegraphics[width=0.2375\columnwidth]{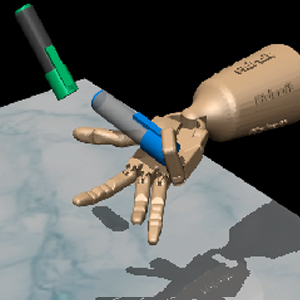} 
  \includegraphics[width=0.2375\columnwidth]{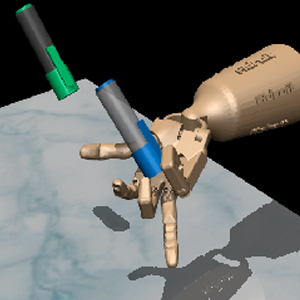} 
  \includegraphics[width=0.2375\columnwidth]{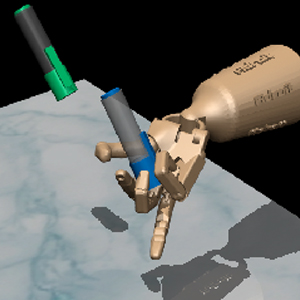} 
  \footnotesize\caption{In-hand manipulation -- reposition the blue pen to match the orientation of the green target. The base of the hand is fixed. The target is randomized to cover all configurations. Task is considered successful when the orientations match within tolerance.}
  \label{fig:pen}
\end{figure}

\begin{figure}[t!]
  \centering
  \includegraphics[width=0.2375\columnwidth]{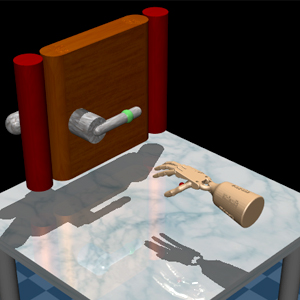} 
  \includegraphics[width=0.2375\columnwidth]{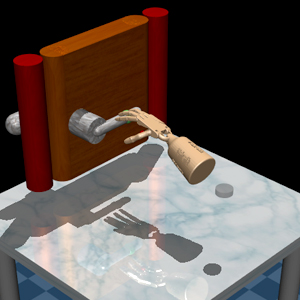} 
  \includegraphics[width=0.2375\columnwidth]{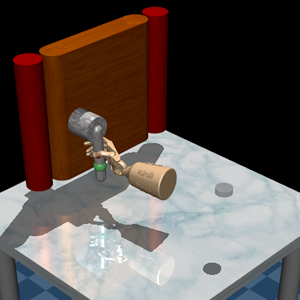} 
  \includegraphics[width=0.2375\columnwidth]{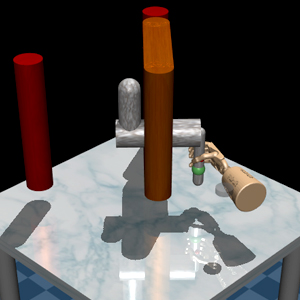} 
  \footnotesize\caption{Door opening -- undo the latch and swing the door open. The latch has significant dry friction and a bias torque that forces the door to stay closed. Agent leverages environmental interaction to develop the understanding of the latch as no information about the latch is explicitly provided. The position of the door is randomized. Task is considered complete when the door touches the door stopper at the other end.}
  \label{fig:doorHandle}
\end{figure}

\begin{figure}[t!]
  \centering
  \includegraphics[width=0.2375\columnwidth]{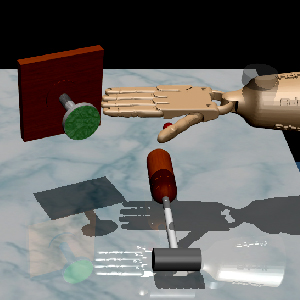} 
  \includegraphics[width=0.2375\columnwidth]{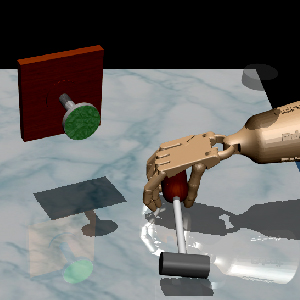} 
  \includegraphics[width=0.2375\columnwidth]{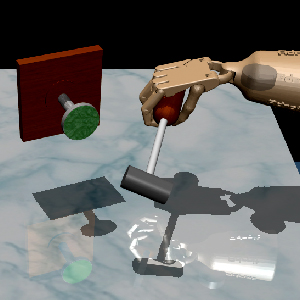} 
   \includegraphics[width=0.2375\columnwidth]{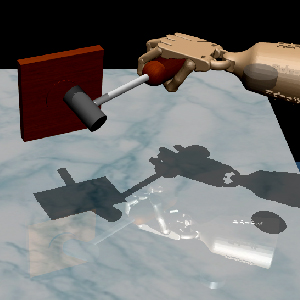} 
  \footnotesize\caption{Tool use -- pick up and hammer with significant force to drive the nail into the board.  Nail position is randomized and has dry friction capable of absorbing up to 15N force. Task is successful when the entire length of the nail is inside the board.}
\label{fig:handle}
\end{figure}

\subsection{Tasks}
\subsubsection{\textbf{Object relocation} (Figure \ref{fig:pickup})} 
Object relocation is a major class of problems in dexterous manipulation, where an object is picked up and moved to a target location. The principal challenge here from an RL perspective is exploration, since in order to achieve success, the hand has to reach the object, grasp it, and take it to the target position -- a feat that is very hard to accomplish without priors in the form of shaped rewards or demonstrations.

\subsubsection{\textbf{In-hand Manipulation} -- Repositioning a pen (Figure \ref{fig:pen})} 
In hand-manipulation maneuvers like re-grasping, re-positioning, twirling objects etc. involve leveraging the dexterity of a high DOF manipulator to effectively navigate a difficult landscape filled with constraints and discontinuities imposed by joint limits and frequently changing contacts. Due to the large number of contacts, conventional model-based approaches which rely on accurate estimates of gradients through the dynamics model struggle in these problem settings. The major challenge in these tasks is representing the complex solutions needed for different maneuvers. For these reason, sampling based DRL methods with rich neural network function approximators are particularly well suited for this class of problems. Previous work on in-hand manipulation with RL~\cite{kumar2016optimal} has considered simpler tasks such as twirling a cylinder, but our tasks involve omni-directional repositioning which involves significantly more contact use. Collecting human demonstrations for this task was challenging due to lack of tactile feedback in VR. Instead, to illustrate the effectiveness of our proposed algorithms, we used a computational expert trained using RL on a well shaped reward for many iterations. This expert serves to give demonstrations which are used to speed up training from scratch.

\subsubsection{\textbf{Manipulating Environmental Props} (Figure \ref{fig:doorHandle})} 
Real-world robotic agents will require constant interaction and manipulation in human-centric environments. Tasks in this class involve modification of the environment itself - opening drawers for fetching, moving furniture for cleaning, etc. The solution is often multi-step with hidden subgoals (e.g undo latch before opening doors), and lies on a narrow constrained manifold shaped primarily by the inertial properties and the under actuated dynamics of the environment.

\subsubsection{\textbf{Tool Use} -- Hammer (Figure \ref{fig:handle})} 
Humans use tools such as hammers, levers, etc. to augment their capabilities. These tasks involve co-ordination between the fingers and the arm to apply the tool correctly. Unlike object relocation, the goal in this class of tasks is to use the tool as opposed to just relocating it. Not all successful grasp leads to effective tool use. Effective tool use requires multiple steps involving grasp reconfiguration and careful motor co-ordination in order to impart the required forces. In addition, effective strategies needs to exhibit robust behaviors in order to counter and recover from destabilizing responses from the environment.

Further details about the tasks, including detailed shaped reward functions, physics parameters etc. are available in the project website: \url{http://sites.google.com/view/deeprl-dexterous-manipulation}.

\subsection{Experimental setup}
\label{sec:experimental_setup}

To accomplish the tasks laid out above, we use a high degree of freedom dexterous manipulator and a virtual reality demonstration system which we describe below. 

\subsubsection{\bf ADROIT hand} 
\begin{figure}[!t]
  \centering
  \includegraphics[angle=90,origin=c,width= 0.5\columnwidth]{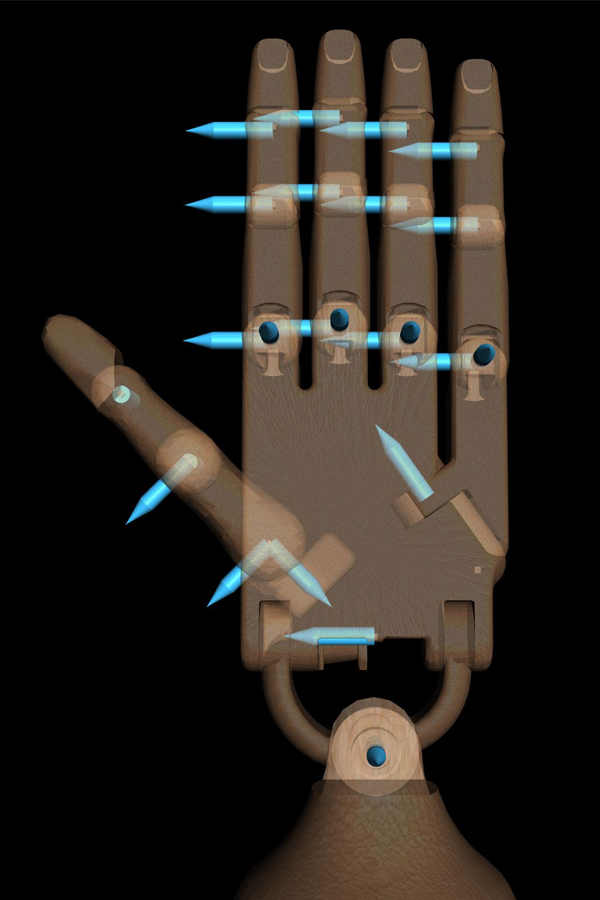}
  \vspace{-2em}
  \caption{24 degree of freedom ADROIT hand. The blue arrows mark the position of the joints and corresponding position actuator.}
  \label{fig:Adroit}
\end{figure}

We use a simulated analogue of a highly dexterous manipulator -- ADROIT~\cite{kumar2013fast}, which is a 24-DoF anthropomorphic platform designed for addressing challenges in dynamic and dexterous manipulation~\cite{kumar2016optimal,kumar2014real}. The first, middle, and ring fingers have 4 DoF. Little finger and thumb have 5 DoF, while the wrist has 2 DoF. Each DoF is actuated using position control  and is equipped with a joint angle sensor (Figure \ref{fig:Adroit}). 

\subsubsection{\bf Simulator} Our experimental setup uses the MuJoCo physics simulator~\cite{Todorov12}. The stable contact dynamics of MuJoCo~\cite{erez2015simulation} makes it well suited for contact rich hand manipulation tasks. The kinematics, the dynamics, and the sensing details of the physical hardware were carefully modeled to encourage physical realism. In addition to dry friction in the joints, all hand-object contacts have planar friction. Object-fingertip contacts support torsion and rolling friction. Though the simulation supports tactile feedback, we do not use it in this work for simplicity, but expect that its use will likely improve the performance.

\subsection{Demonstrations}
\label{sec:demo_collection}
Accurate demonstrations data are required to help accelerate various learning algorithms. Standard methods like kinesthetic teaching are impractical with complex systems like ones we study in this work. We use an updated version of the Mujoco HAPTIX system~\cite{kumar2015mujoco}. The system uses the CyberGlove III system for recording the fingers, HTC vive tracker for tracking the base of the hand and HTC vive headset for stereoscopic visualization.
This moves the process of demonstration data collection from the real world to virtual reality, allowing for several high fidelity demonstrations for tasks involving large number of contacts and dynamic phenomena such as rolling, sliding, stick-slip, deformations and soft contacts. Since the demonstrations are provided in simulation, physically consistent details of the movements can be easily recorded. We gathered 25 successful demonstrations for all our tasks (with task randomization as outlined in captions of Figure \ref{fig:pickup}, \ref{fig:pen}, \ref{fig:handle}, and \ref{fig:doorHandle}), with each demonstration consisting of the state-action trajectories needed to perform the task in simulation. To combat distribution drift, a small amount of noise (uniform random $[-0.1, 0.1]$ radians) is added to the actuators per timestep so that the policy can better capture relevant statistics about the data.

\section{Demo Augmented Policy Gradient (DAPG)}
In this work, we use a combination of RL and imitation learning to solve complex dexterous manipulation problems. To reduce sample complexity and help with exploration, we collect a few expert demonstrations using the VR system described in Section~\ref{sec:demo_collection}, and incorporate these into the RL process. We first present some RL preliminaries, followed by the base RL algorithm we use for learning, and finally describe our procedure to incorporate demonstrations. 

\subsection{Preliminaries}
We model the control problem as a Markov decision process (MDP), which is defined using the tuple: \hbox{$\cM = \lbrace \cS, \cA, \cR, \cT, \rho_0, \gamma \rbrace$}. $\cS \in \bR^n$ and $\cA \in \bR^m$ represent the state and actions. \hbox{$\cR: \cS \times \cA \rightarrow \bR$} is the reward function which measures task progress. In the ideal case, this function is simply an indicator function for task completion (i.e. a {\it sparse task completion reward}). In practice, various forms of {\it shaped rewards} that incorporate human priors on how to accomplish the task might be required to make progress on the learning problem. $\cT: \cS \times \cA \rightarrow \cS$ is the transition dynamics, which can be stochastic. In model-free RL, we do not assume knowledge about this transition function, and require only sampling access to this function. $\rho_0$ is the probability distribution over initial states and $\gamma \in [0,1)$ is a discount factor. We denote the demonstrations data set using $\rho_D = \Big\{ \left(s_t^{(i)}, a_t^{(i)}, s_{t+1}^{(i)}, r_t^{(i)} \right) \Big\}$, where $t$ indexes time and $i$ indexes different trajectories. These demonstrations can be used to guide reinforcement learning and significantly reduce sample complexity of RL. We wish to solve for a stochastic policy of the form \hbox{$\pi: \cS \times \cA \rightarrow \bR_+$}, which optimizes the expected sum of rewards:
\begin{equation}
\eta(\pi) = \bE_{\pi, \cM} \Bigg[ \sum_{t=0}^\infty \gamma^t r_t \Bigg].
\end{equation}
We also define the value, $Q$, and advantage functions:
$$ \Vpi(s) = \bE_{\pi, \cM} \Bigg[ \sum_{t=0}^\infty \gamma^t r_t \mid s_0 = s \Bigg] $$ 
$$ \Qpi(s,a) = \bE_{\cM} \Big[\cR(s,a) \Big] + \bE_{s'\sim \cT(s,a)} \Big[ \Vpi(s') \Big] $$
$$ \Api(s,a) = \Qpi(s,a) - \Vpi(s). $$
We consider parameterized policies $\pith$, and hence wish to optimize for the parameters $(\theta)$. Thus, we overload notation and use $\eta(\pi)$ and $\eta(\theta)$ interchangeably.

\subsection{Natural Policy Gradient}
In this work, we primarily consider policy gradient methods, which are a class of model-free RL methods. In policy gradient methods, the parameters of the policy are directly optimized to maximize the objective, $\eta(\theta)$, using local search methods such as gradient ascent. In particular, for this work we consider the NPG algorithm~\cite{Kakade01,Peters07,Rajeswaran17}. First, NPG computes the vanilla policy gradient, or REINFORCE~\cite{Williams92} gradient:
\begin{equation}
g = \frac{1}{NT} \ \sum_{i=1}^{N} \sum_{t=1}^T \nabla_\theta \log \pith(a^i_t|s^i_t) \hat{\Api}(s^i_t, a^i_t, t).
\end{equation}
Secondly, it pre-conditions this gradient with the (inverse of) Fisher Information Matrix~\cite{Amari98,Kakade01} computed as:
\begin{equation}
F_{\theta} = \frac{1}{NT} \ \sum_{i=1}^{N} \nabla_\theta \log \pith(a^i_t|s^i_t) \nabla_\theta \log \pith(a^i_t|s^i_t)^T,
\end{equation}
and finally makes the following normalized gradient ascent update~\cite{Peters07,Schulman15,Rajeswaran17}:
\begin{equation}
\theta_{k+1} = \theta_k + \sqrt{\frac{\delta}{g^T F_{\theta_k}^{-1}g }} \ F_{\theta_k}^{-1}g,
\end{equation}
where $\delta$ is the step size choice. A number of pre-conditioned policy gradient methods have been developed in literature~\cite{Kakade01,bagnell,Peters,Peters07,Schulman15,Rajeswaran17,ppo} and in principle any of them could be used. Our implementation of NPG for the experiments is based on Rajeswaran et al.~\cite{Rajeswaran17}.

\subsection{Augmenting RL with demonstrations}
\label{sec:demoPG}

Although NPG with an appropriately shaped reward can somewhat solve the tasks we consider, there are several challenges which necessitate the incorporations of demonstrations to improve RL:

\begin{enumerate}
	\item RL is only able to solve the tasks we consider with careful, laborious task reward shaping. 
    \item While RL eventually solves the task with appropriate shaping, it requires an impractical number of samples to learn - in the order of a 100 hours for some tasks. 
    \item The behaviors learned by pure RL have unnatural appearance, are noisy and are not as robust to environmental variations.
\end{enumerate}

Combining demonstrations with RL can help combat all of these issues. Demonstrations help alleviate the need for laborious reward shaping, help guide exploration and decrease sample complexity of RL, while also helping produce robust and natural looking behaviors. We propose the demonstration augmented policy gradient (DAPG) method which incorporates demonstrations into policy gradients in two ways:

\subsubsection{Pretraining with behavior cloning}
Policy gradient methods typically perform exploration by utilizing the stochasticity of the action distribution defined by the policy itself. If the policy is not initialized well, the learning process could be very slow with the algorithm exploring state-action spaces that are not task relevant. To combat this, we use behavior cloning (BC)~\cite{ALVINN, nvidiaimitation} to provide an informed policy initialization that efficiently guides exploration. Use of demonstrations circumvents the need for reward shaping often used to guide exploration. This idea of pretraining with demonstrations has been used successfully in prior work~\cite{DMP2008}, and we show that this can dramatically reduce the sample complexity for dexterous manipulation tasks as well.
BC corresponds to solving the following maximum-likelihood problem:
\begin{equation}
\underset{\theta}{\text{maximize}} \sum_{(s,a) \in \rho_D} \ln \pith(a|s).
\end{equation}
The optimizer of the above objective, called the behavior cloned policy, attempts to mimic the actions taken in the demonstrations at states visited in the demonstrations. In practice, behavior cloning does not guarantee that the cloned policy will be effective, due to the distributional shift between the demonstrated states and the policy's own states~\cite{dagger}. Indeed, we observed experimentally that the cloned policies themselves were usually not successful.

\subsubsection{RL fine-tuning with augmented loss}
Though behavior cloning provides a good initialization for RL, it does not optimally use the information present in the demonstration data. Different parts of the demonstration data are useful in different stages of learning, especially for tasks involving a sequence of behaviors. For example, the hammering task requires behaviors such as reaching, grasping, and hammering. Behavior cloning by itself cannot learn a policy that exhibits all these behaviors in the correct sequence with limited data. The result is that behavior cloning produces a policy that can often pick up the hammer but seldom swing it close to the nail. The demonstration data contains valuable information on how to hit the nail, but is lost when the data is used only for initialization. Once RL has learned to pick up the hammer properly, we should use the demonstration data to provide guidance on how to hit the nail. To capture all information present in the demonstration data, we add an additional term to the gradient:
\begin{equation}
\begin{aligned}
g_{aug} = & \sum_{(s,a) \in \rho_\pi} \nabla_\theta \ln \pith(a|s) A^\pi(s,a) + \\
& \sum_{(s,a) \in \rho_D} \nabla_\theta \ln \pith(a|s) w(s,a).
\end{aligned}
\end{equation}
Here $\rho_\pi$ represents the dataset obtained by executing policy $\pi$ on the MDP, and $w(s,a)$ is a weighting function. This augmented gradient is then used in eq.~(4) to perform a co-variant update. If $w(s,a)=0 \ \forall (s,a)$, then we recover the policy gradient in eq. (2). If $w(s,a)=c \ \forall (s,a)$, with sufficiently large $c$, it reduces to behavior cloning, as in eq.~(5). However, we wish to use both imitation and reinforcement learning, so we require an alternate weighting function. The analysis in~\cite{cpi} suggests that eq.~(2) is also valid for mixture trajectory distributions of the form \hbox{$\rho = \alpha \rho_\pi + (1-\alpha) \rho_D$}. Thus, a natural choice for the weighting function would be \hbox{$w(s,a) = A^\pi(s,a) \ \forall (s,a) \in \rho_D$.} However, it is not possible to compute this quantity without additional rollouts or assumptions~\cite{aggrevated}. Thus, we use the heuristic weighting scheme: 
$$w(s,a) = \lambda_0 \lambda_1^k \max_{(s',a') \in \rho_\pi} A^\pi(s',a') \ \ \forall (s,a) \in \rho_D, $$ 
where $\lambda_0$ and $\lambda_1$ are hyperparameters, and $k$ is the iteration counter. The decay of the weighting term via $\lambda_1^k$ is motivated by the premise that initially the actions suggested by the demonstrations are at least as good as the actions produced by the policy. However, towards the end when the policy is comparable in performance to the demonstrations, we do not wish to bias the gradient. Thus, we asymptotically decay the auxiliary objective. We empirically find that the performance of the algorithm is not very sensitive to the choice of these hyperparameters. For all the experiments, $\lambda_0=0.1$ and $\lambda_1=0.95$ was used.

\section{Results and Discussion} 
\label{sec:results}

\begin{figure*}
    \centering
    \includegraphics[trim={0cm, 0, 0cm, 0}, clip, width=0.265\textwidth]{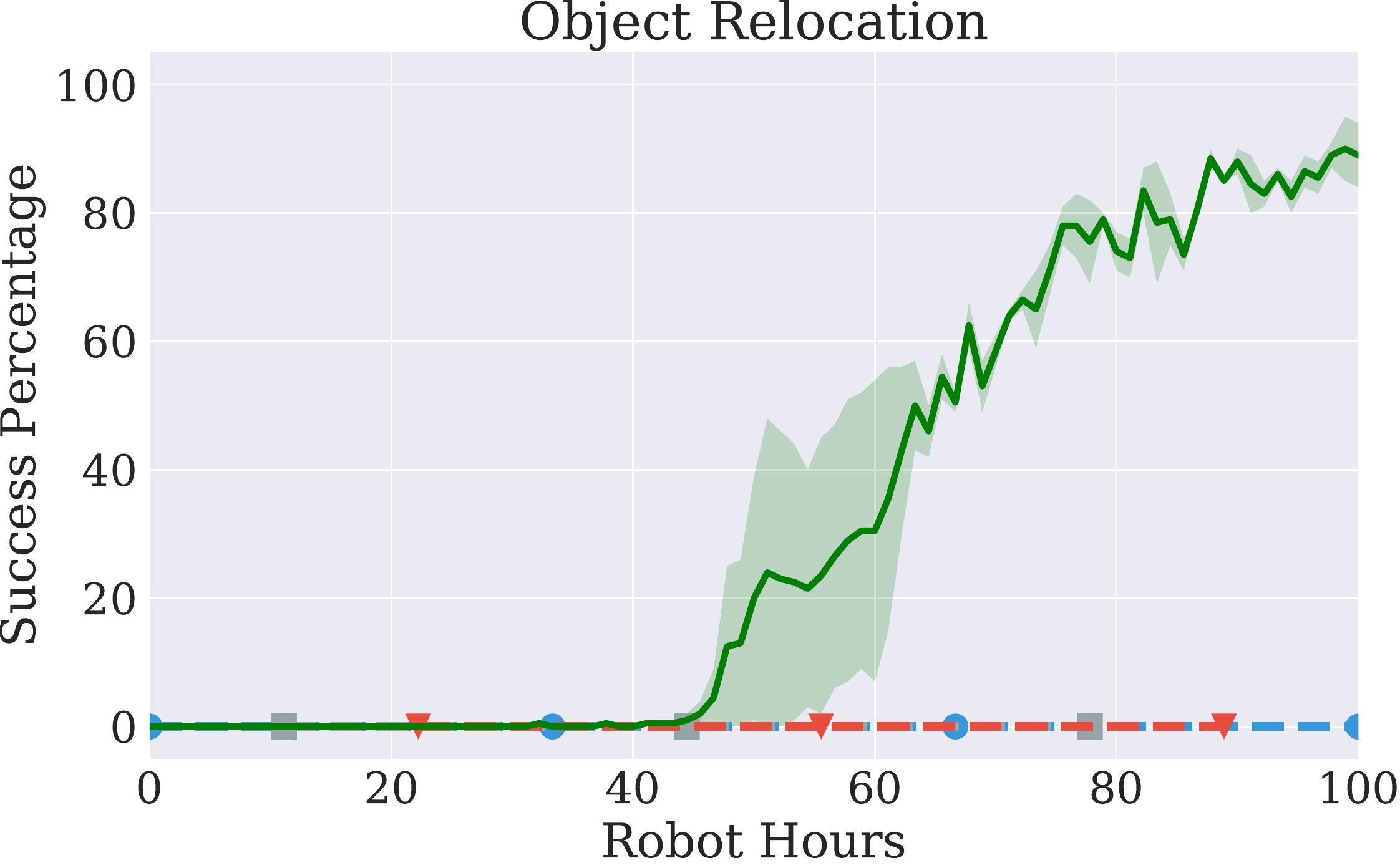}
    \includegraphics[trim={2.2cm, 0, 0cm, 0}, clip, width=0.2375\textwidth]{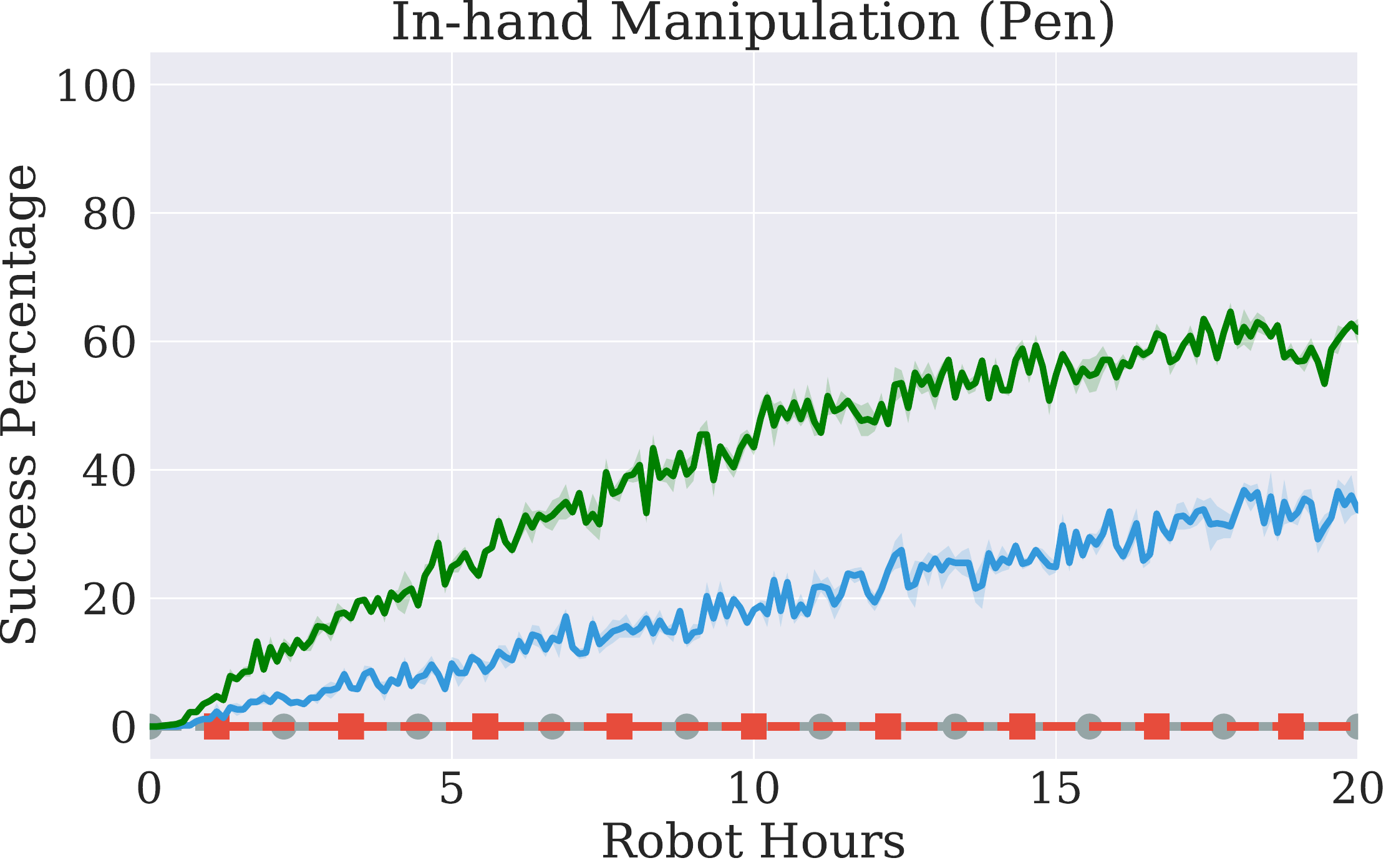}
    \includegraphics[trim={2.2cm, 0, 0cm, 0}, clip, width=0.2375\textwidth]{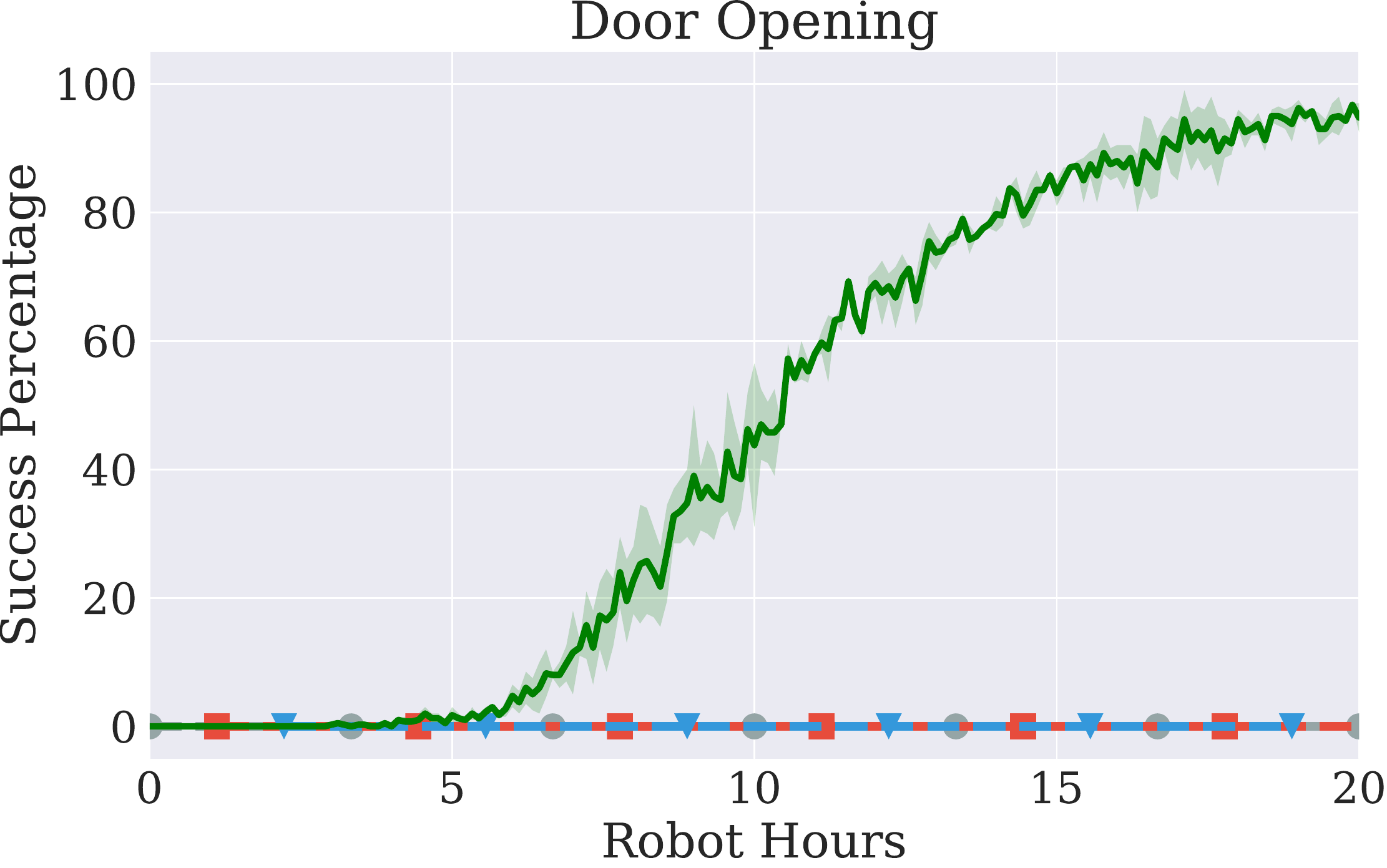}
    \includegraphics[trim={2.2cm, 0, 0cm, 0}, clip, width=0.2375\textwidth]{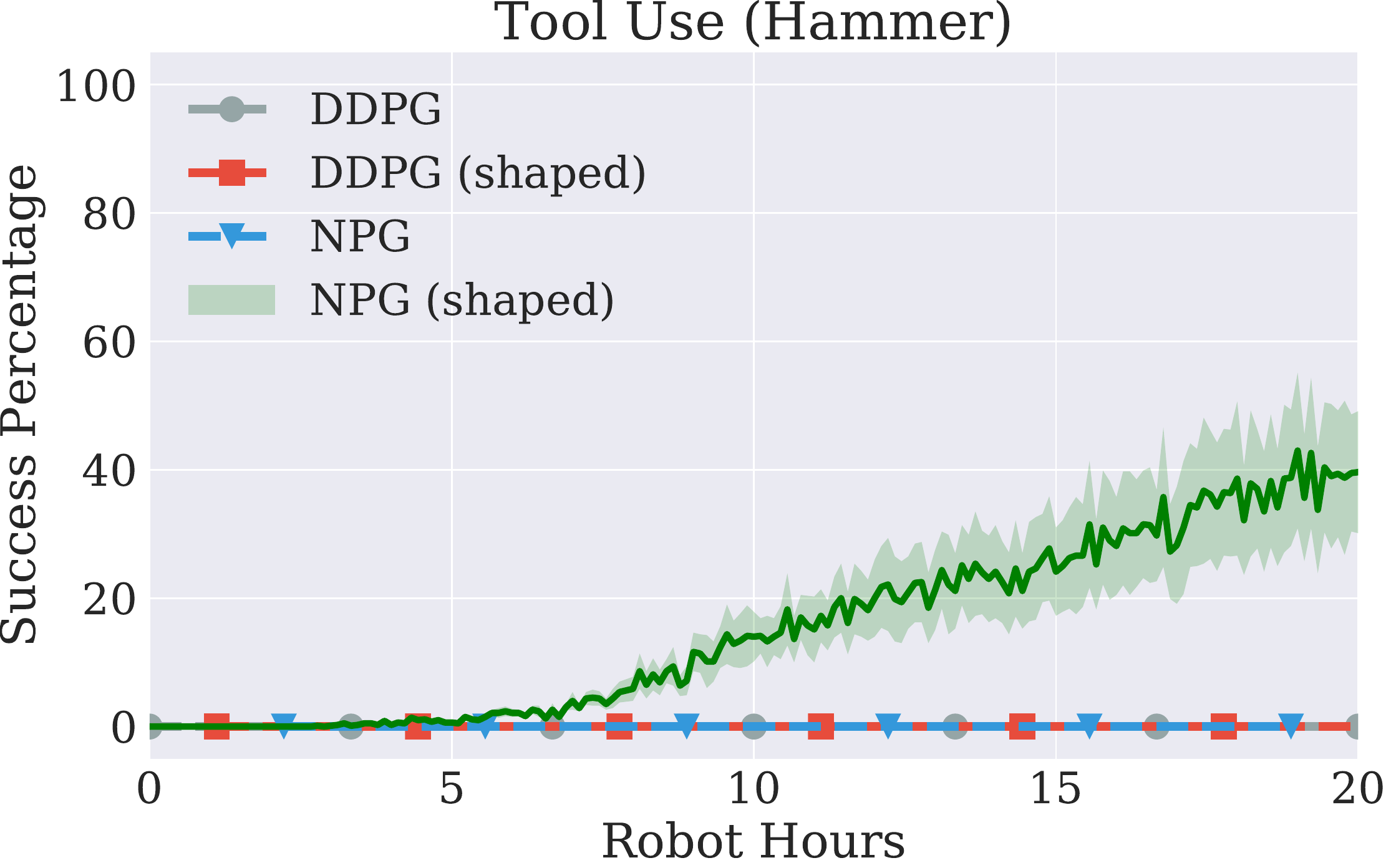}
   \caption{Performance of pure RL methods -- NPG and DDPG, with sparse task completion reward and shaped reward. Sparse reward setting is primarily ineffective in solving our task set (expect in-hand manipulation). Incorporating human priors using reward shaping helps NPG get off the ground, but DDPG sill struggles to find success.}
    \label{fig:pureRL}
   \vspace{-1em}
\end{figure*} 

Our results study how RL methods can learn dexterous manipulation skills, comparing several recent algorithms and reward conditions. First, we evaluate the capabilities of RL algorithms to learn dexterous manipulation behaviors from scratch on the tasks outlined in Section~III. Subsequently, we demonstrate the benefits of incorporating human demonstrations with regard to faster learning, increased robustness of trained policies, and ability to cope with sparse task completion rewards.

\subsection{Reinforcement Learning from Scratch} 
We aim to address the following questions in this experimental evaluation:
\begin{enumerate}
\item Can existing RL methods cope with the challenges presented by the high dimensional dexterous manipulation tasks?
\item Do the resulting policies exhibit desirable properties like robustness to variations in the environment?
\item Are the resulting movements safe for execution on physical hardware, and are elegant/nimble/human-like?
\end{enumerate}

 \begin{figure}[h]
  \centering
  \includegraphics[width=0.2375\columnwidth]{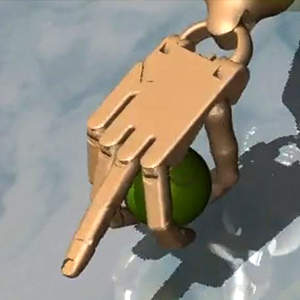} 
  \includegraphics[width=0.2375\columnwidth]{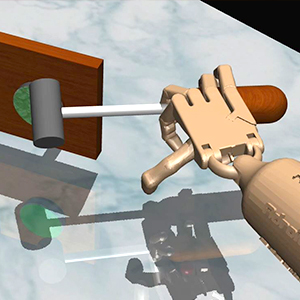} 
  \includegraphics[width=0.2375\columnwidth]{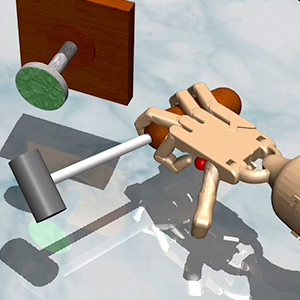} 
  \includegraphics[width=0.2375\columnwidth]{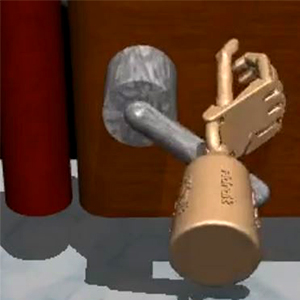} 
  \footnotesize\caption{Unnatural movements observed in the execution trace of behavior trained with pure reinforcement leaning. From left to right: (a) unnatural, socially unacceptable, finger position during pick up. (b/c) unnatural grasp for hammer (d) unnatural use of wrist for unlatching the door.}
  \label{fig:funnyMov}
   \vspace{-2em}
\end{figure}

In order to benchmark the capabilities of DRL with regard to the dexterous manipulation tasks outlined in Section~\ref{sec:tasks}, we evaluate the NPG algorithm described briefly in Section~V, and the DDPG algorithm~\cite{ddpg}, which has recently been used in a number of robotic manipulation scenarios~\cite{DDPGfD,shaneNAF,ashvin}. Both of these methods have demonstrated state of the art results in popular DRL continuous control benchmarks, and hence serve as a good representative set. We score the different methods based on the percentage of successful trajectories the trained policies can generate, using a sample size of 100 trajectories. We find that with sparse task completion reward signals, the policies with random exploration never experience success (except in the in-hand task) and hence do not learn.

In order to enable these algorithms to learn, we incorporate human priors on how to accomplish the task through careful \textit{reward shaping}. With the shaped rewards, we find that NPG is indeed able to achieve high success percentage on these tasks (Figure~\ref{fig:pureRL}), while DDPG was unable to learn successful policies despite considerable hyperparameter tuning. DDPG can be very sample efficient, but is known to be very sensitive to hyperparameters and random seeds~\cite{deeprlthatmatters}, which may explain the difficulty of scaling it to complex, high-dimensional tasks like dexterous manipulation.

\begin{figure}[h]
  \centering
  \includegraphics[width=0.45\columnwidth]{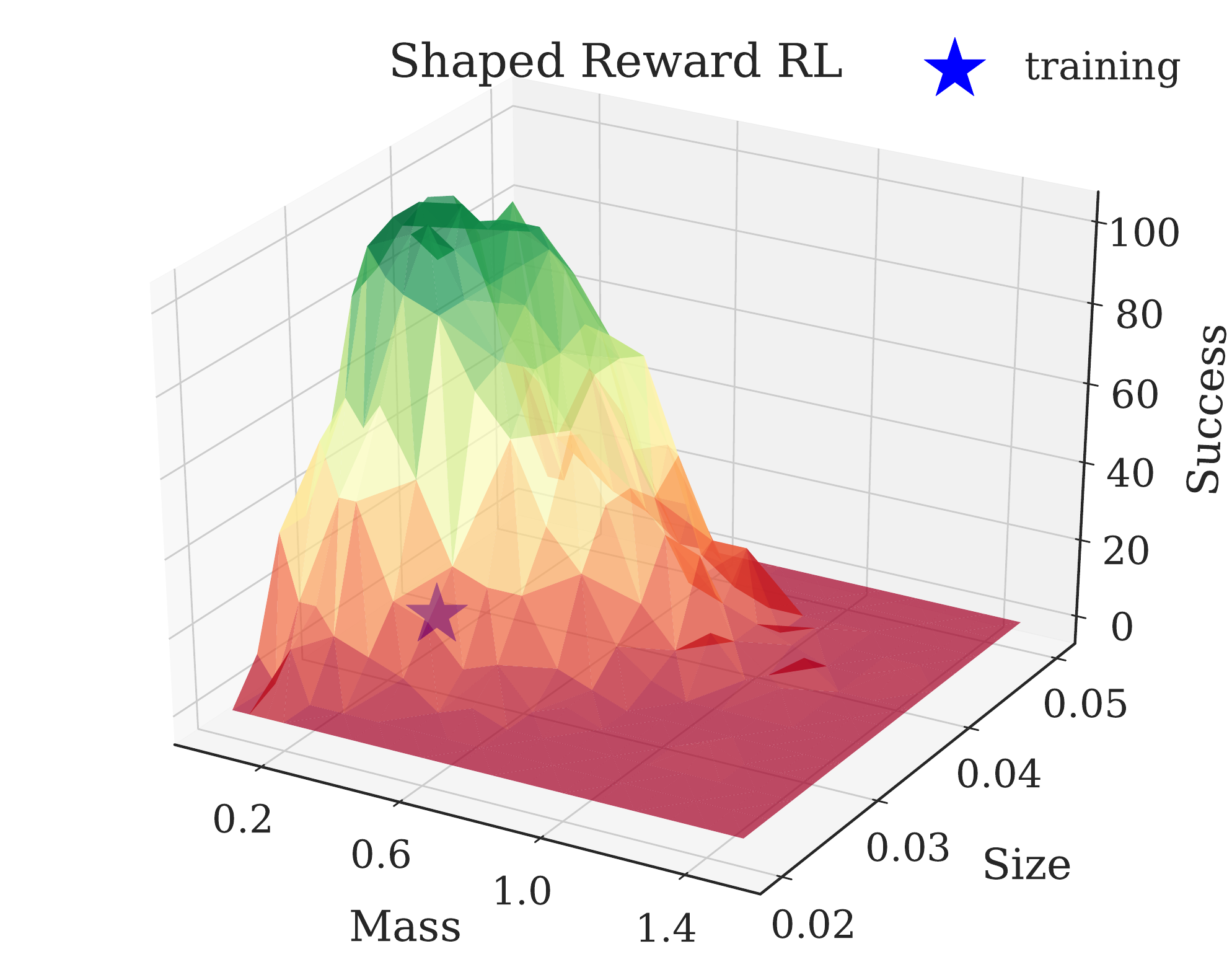} 
  \includegraphics[width=0.45\columnwidth]{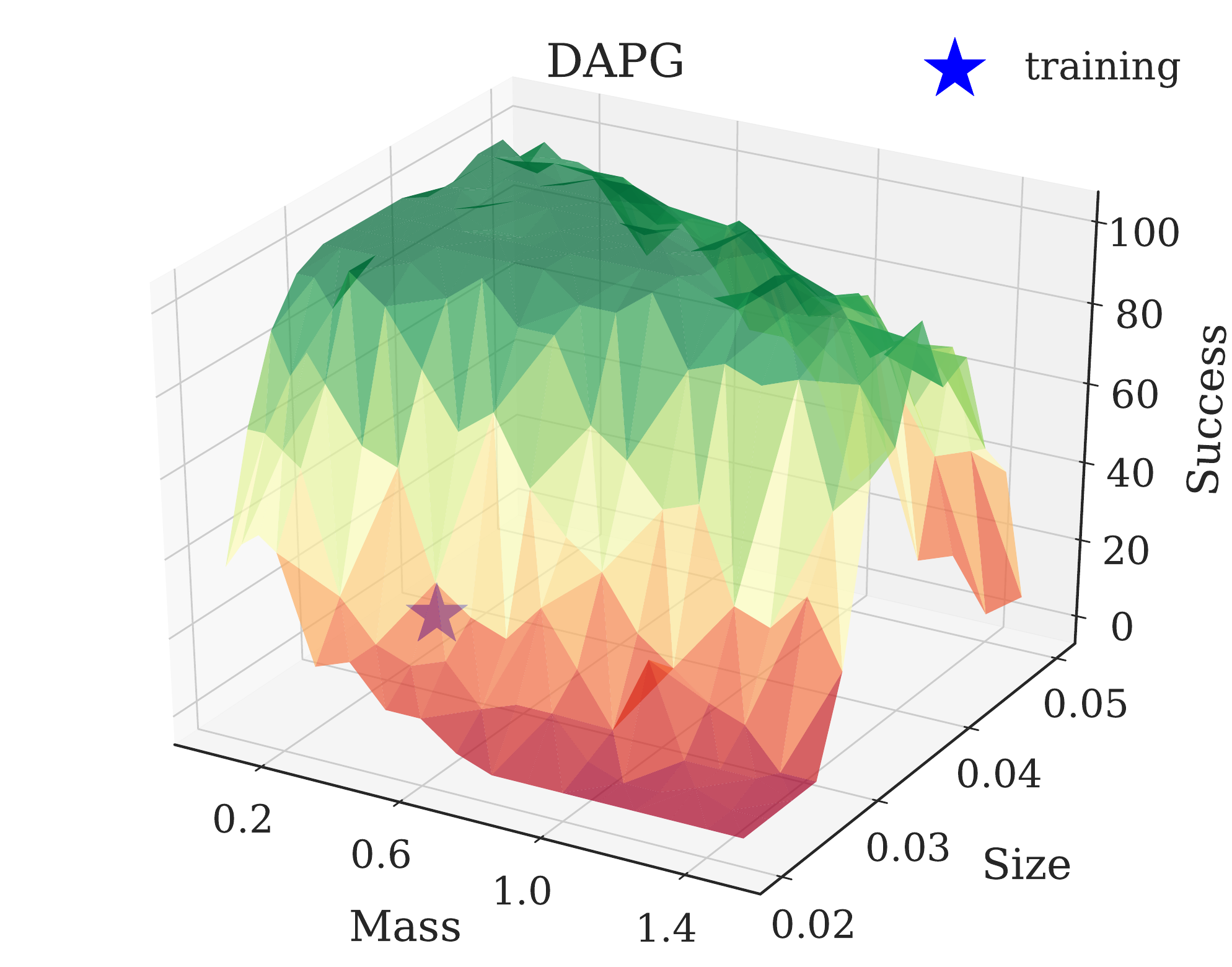} \\
  \includegraphics[width=0.45\columnwidth]{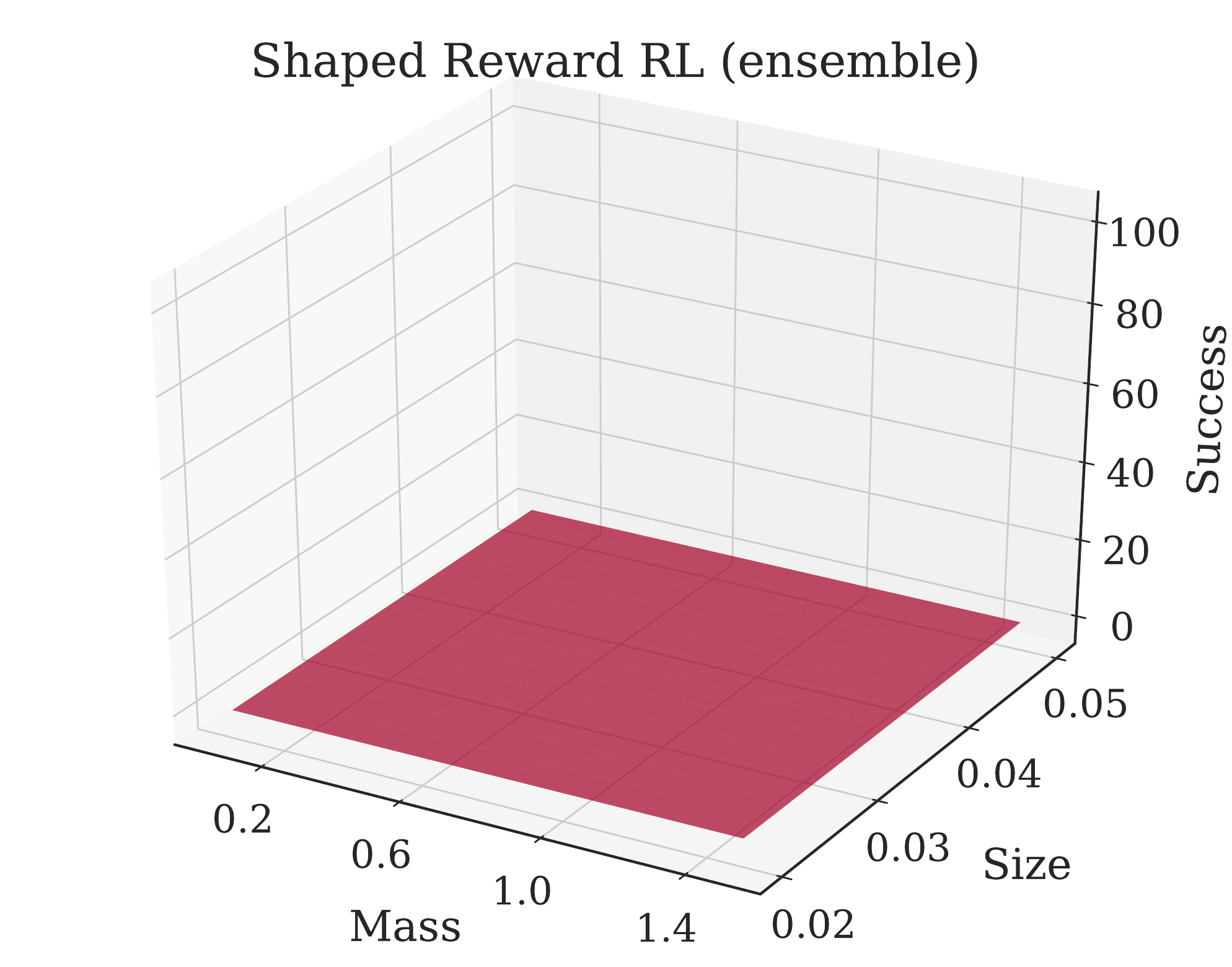}
  \includegraphics[width=0.45\columnwidth]{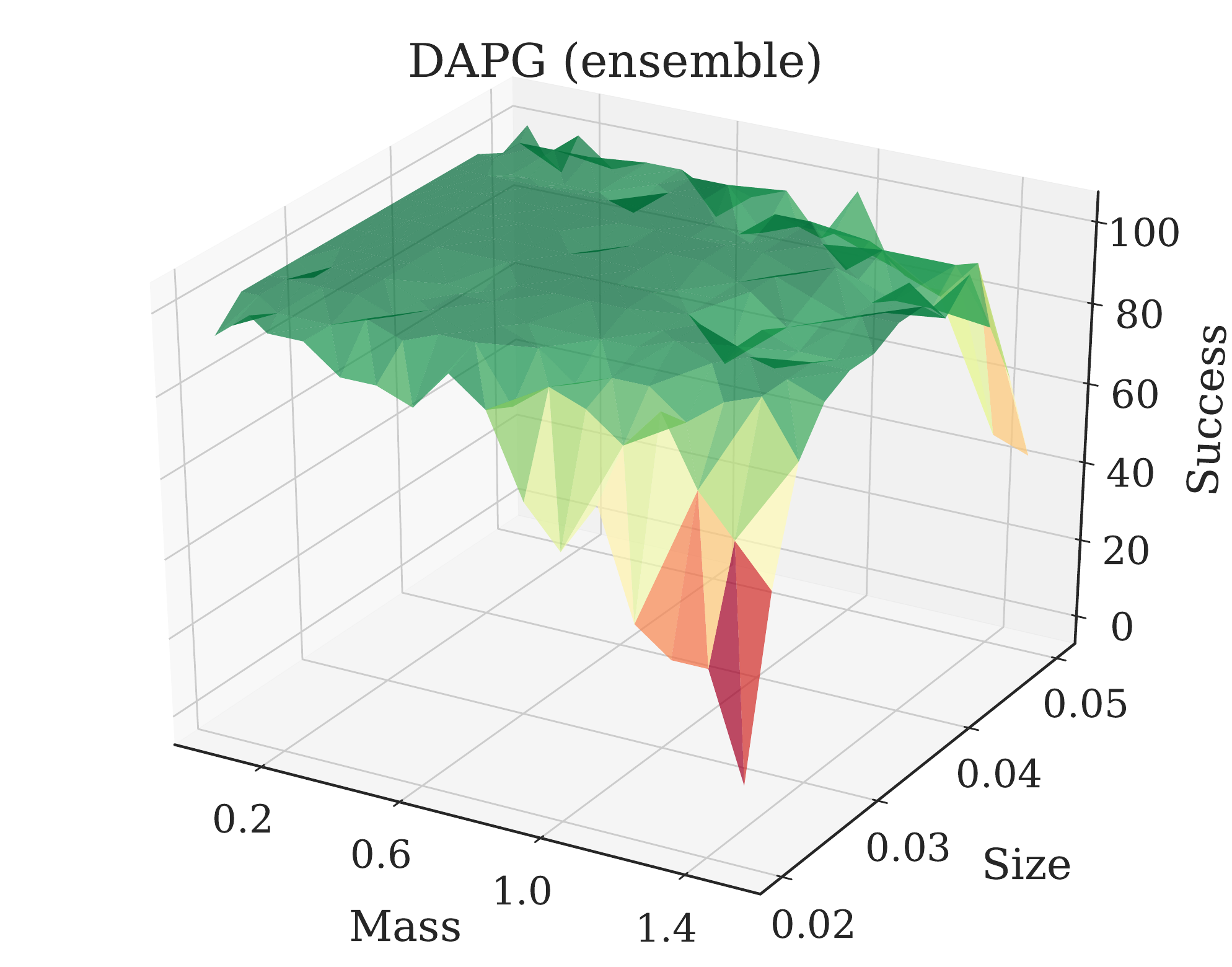}
  \footnotesize\caption{Robustness of trained policies to variations in the environment. The top two figures are trained on a single instance of the environment (indicated by the star) and then tested on different variants of the environment. The policy trained with DAPG is more robust, likely due to the intrinsic robustness of the human strategies which are captured through use of demonstrations. We also see that RL from scratch is unable to learn when the diversity of the environment is increased (bottom left) while DAPG is able to learn and produces even more robust policies.}
	\label{fig:robustness}
\vspace{-2em}
\end{figure}

Although incorporation of human knowledge via reward shaping is helpful, the resulting policies: (a)~often exhibit unnatural looking behaviors, and (b)~are too sample inefficient to be useful for training on the physical hardware. While it is hard to mathematically quantify the quality of generated behaviors, Figure~\ref{fig:funnyMov} and the accompanying video clearly demonstrate that the learned policies produce behaviors that are erratic and not human-like. Such unnatural behaviors are indeed quite prevalent in the recent DRL results~\cite{parkour}. Furthermore, we take the additional step of analyzing the robustness of these policies to variations in environments that were not experienced during training. To do so, we take into account the policy trained for the object relocation task and vary the mass and size of the object that has to be relocated. We find that the policies tend to over-fit to the specific objects they were trained to manipulate and is unable to cope with variations in the environment as seen in Figure~\ref{fig:robustness}.

We attribute the artifacts and brittleness outlined above to the way in which human priors are incorporated into the policy search process. The mental models of solution strategies that humans have for these tasks are indeed quite robust. However it is challenging to distill this mental model or intuition into a mathematical reward function. As we will show in the next section, using a data driven approach to incorporate human priors, in the form of demonstrations, alleviates these issues to a significant extent.

\subsection{Reinforcement Learning with Demonstrations} 
\label{sec:RLwithDemos}

\begin{figure*}[t!]
    \centering
    \includegraphics[trim={0cm, 0, 0cm, 0}, clip, width=0.265\textwidth]{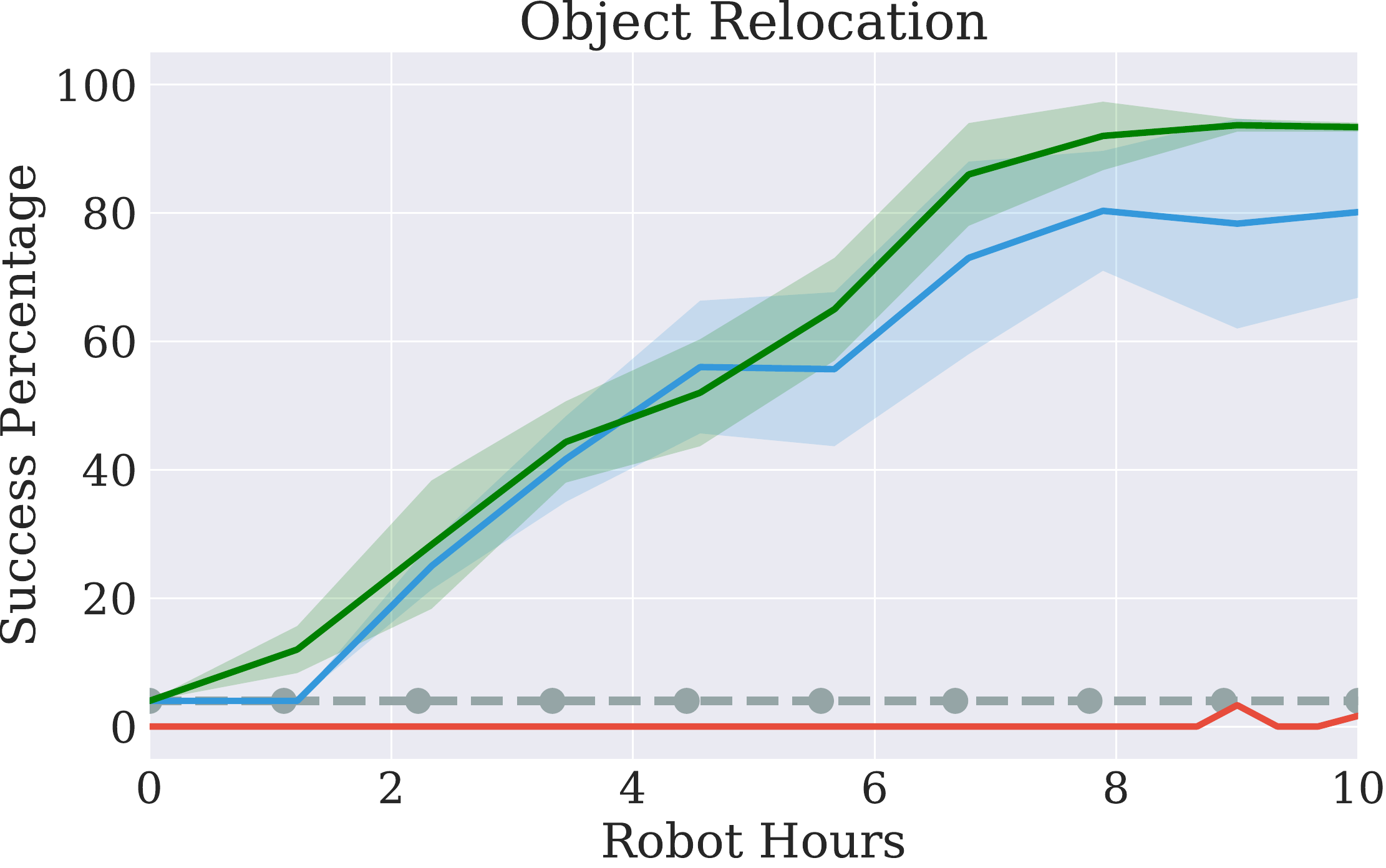}
    \includegraphics[trim={2.2cm, 0, 0cm, 0}, clip, width=0.2375\textwidth]{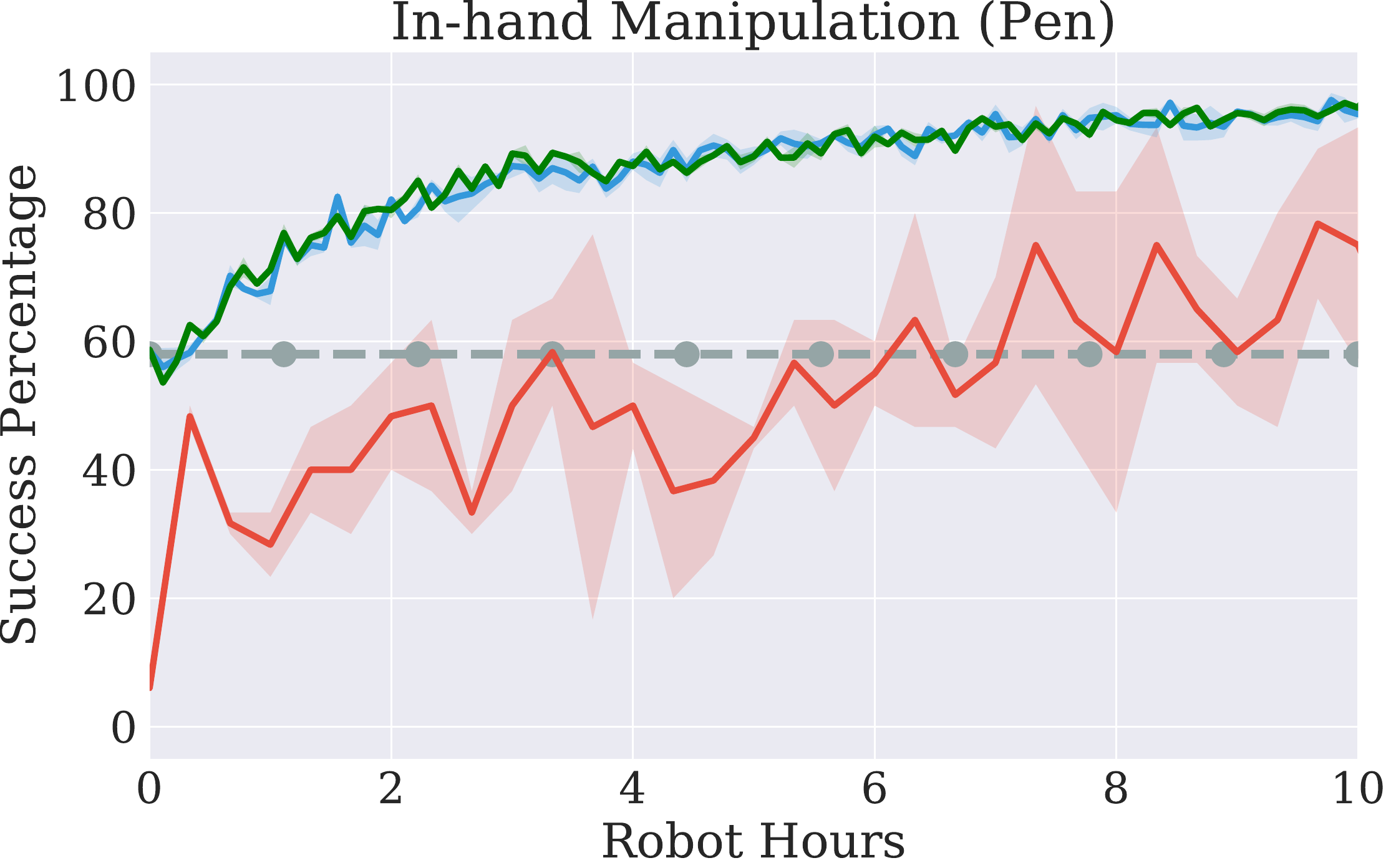}
    \includegraphics[trim={2.2cm, 0, 0cm, 0}, clip, width=0.2375\textwidth]{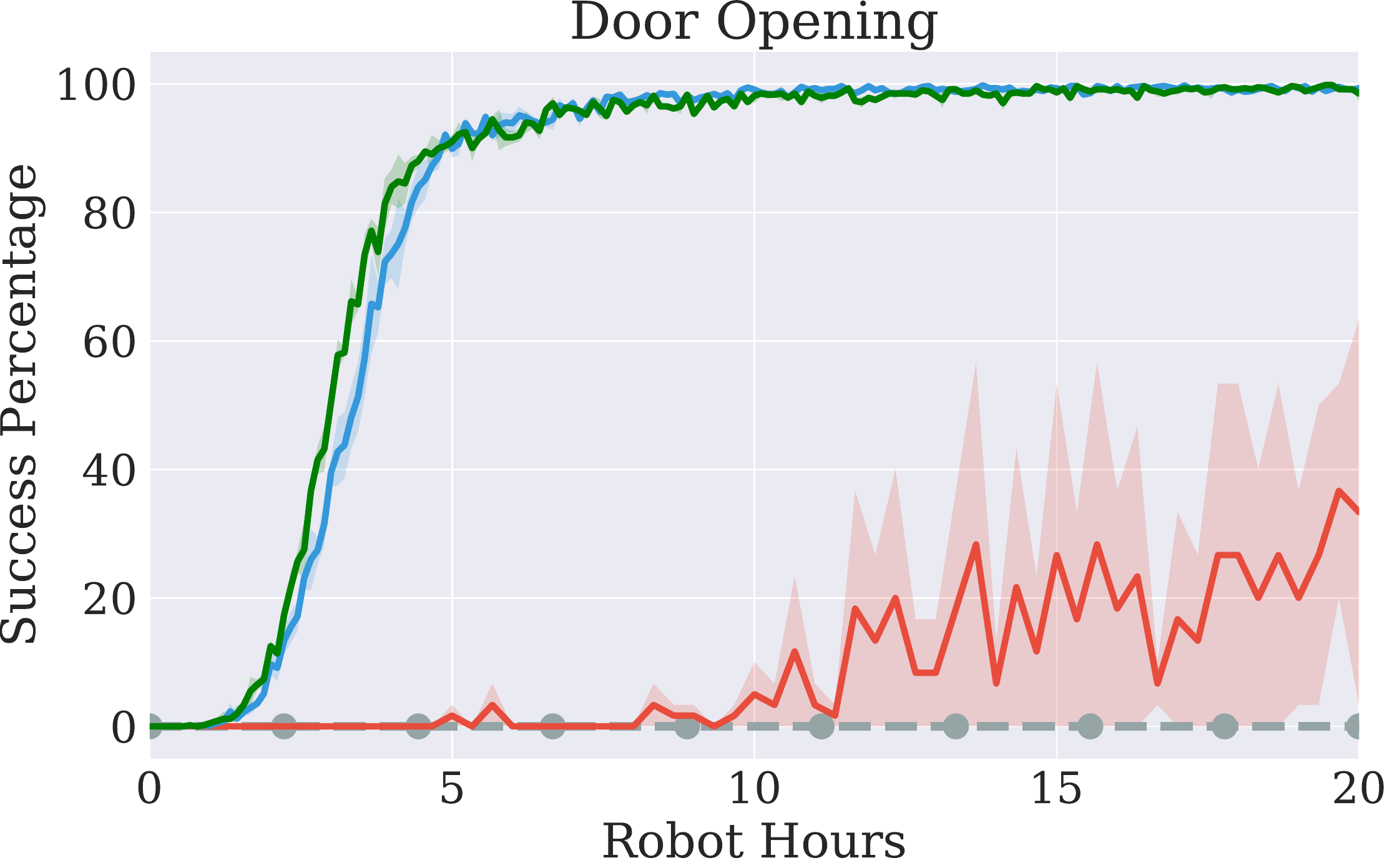}
    \includegraphics[trim={2.2cm, 0, 0cm, 0}, clip, width=0.2375\textwidth]{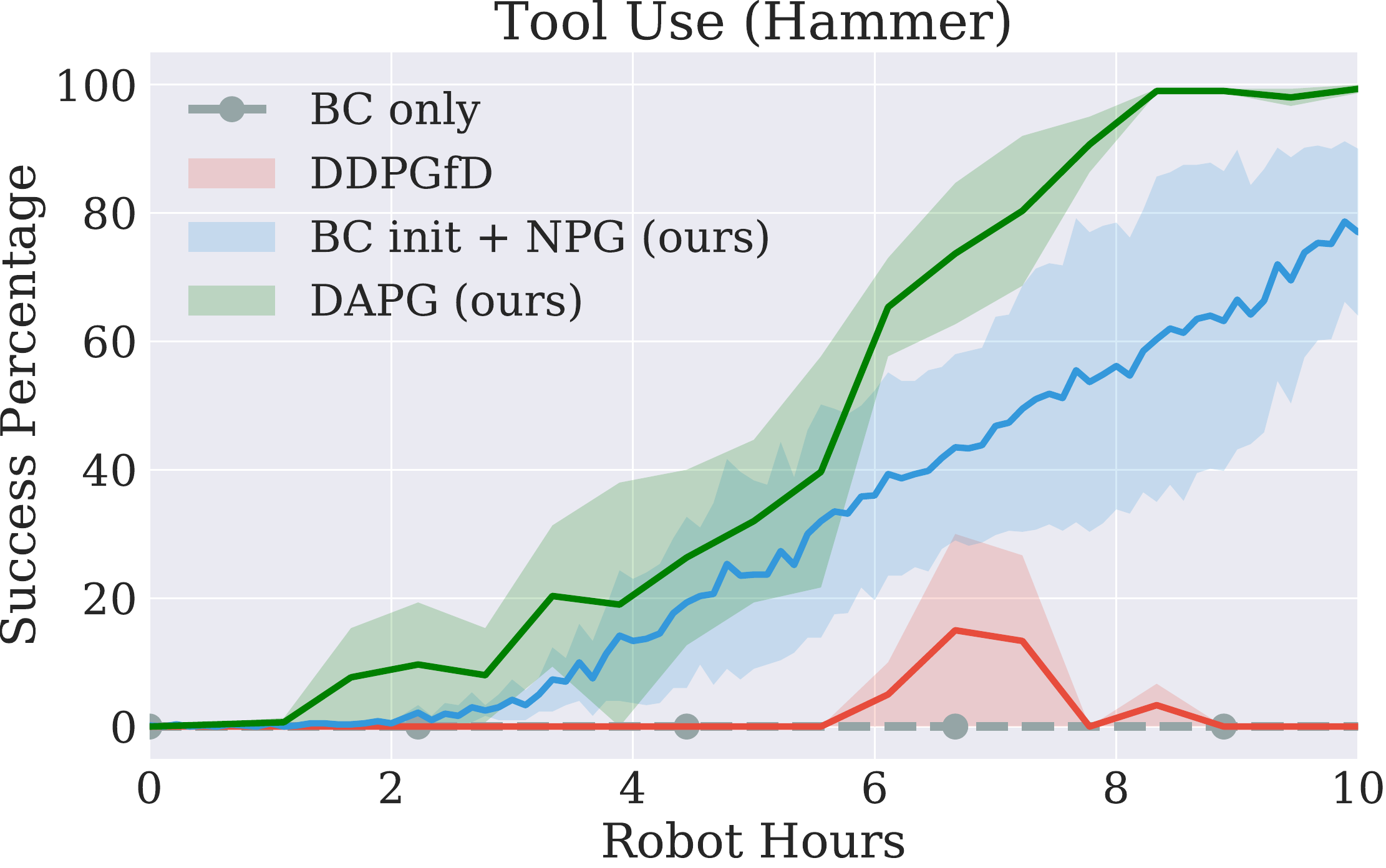}
    \caption{Performance of RL with demonstrations methods -- DAPG(ours) and DDPGfD. DAPG significantly outperforms DDPGfD. For DAPG, we plot the performance of the stochastic policy used for exploration. At any iteration, the performance of the underlying deterministic policy will be better.}
    \label{fig:RLwithDemos}
\end{figure*} 

In this section we aim to study the following questions:
\begin{enumerate}
\item Does incorporating demonstrations via DAPG reduce the learning time to practical timescales?
\item How does DAPG compare to other model-free methods that incorporate demonstrations, such as DDPGfD~\cite{DDPGfD}?
\item Does DAPG acquire robust and human-looking behaviors without reward shaping?
\end{enumerate}

\begin{table}[t]
\centering
\caption{Sample and robot time complexity of DAPG (ours) compared to RL (Natural Policy Gradient) from scratch with shaped (sh) and sparse task completion reward (sp). $N$ is the number of RL iterations needed to achieve $90\%$ success rate, Hours represent the robot hours needed to learn the task. Each iteration is 200 trajectories of length 2 seconds each.}
\label{my-label}
\begin{tabular}{@{}l|ll|ll|ll@{}}
\toprule
Method & DAPG &(sp) & RL  &(sh) & RL &(sp)\\ \midrule
Task             & $N$ & Hours & $N$ & Hours & $N$ & Hours\\ \midrule
Relocation  & 52         & 5.77      & 880         & 98    & $\infty$    & $\infty$    \\
Hammer          & 55            & 6.1       & 448         & 50    & $\infty$    & $\infty$      \\
Door            & 42            & 4.67      & 146         & 16.2  & $\infty$    & $\infty$    \\
Pen             & 30            & 3.33      & 864         & 96    & 2900        & 322   \\ \bottomrule
\end{tabular}
\label{tab:hours}
\vspace{-2em}
\end{table}

We employ the DAPG algorithm in Section~\ref{sec:demoPG} on the set of hand tasks and compare with the recently proposed DDPGfD method~\cite{DDPGfD}. DDPGfD builds on top of the DDPG algorithm, and incorporates demonstrations to bootstrap learning by: (1)~Adding demonstrations to the replay buffer; (2)~Using prioritzed experience replay; (3)~Using n-step returns; (4)~Adding regularizations to the policy and critic networks. Overall, DDPGfD has proven effective on arm manipulation tasks with sparse rewards in prior work, and we compare performance of DAPG against DDPGfD on our dexterous manipulation tasks.

For this section of the evaluation we use only sparse task completion rewards, since we are using demonstrations. With the use of demonstrations, we expect the algorithms to implicitly learn the human priors on how to accomplish the task. Figure~\ref{fig:RLwithDemos} presents the comparison between the different algorithms. DAPG convincingly outperforms DDPGfD in all the tasks well before DDPGfD even starts showing signs of progress. Furthermore, DAPG is able to train policies for these complex tasks in under a few robot hours Table~\ref{tab:hours}. In particular, for the object relocation task, DAPG is able to train policies almost 30 times faster compared to learning from scratch.  This indicates that RL methods in conjunction with demonstrations, and in particular DAPG, are viable approaches for real world training of dexterous manipulation tasks.

When analyzing the robustness of trained policies to variations in the environment, we find that policies trained with DAPG are significantly more robust compared to the policies trained with shaped rewards ({Figure ~\ref{fig:robustness}). Furthermore, a detail that can be well appreciated in the accompanying video is the human-like motions generated by the policy. The policies trained with DAPG better capture the human priors  from the demonstrations, and generate policies that are more robust and exhibit qualities that are inherently expected but hard to mathematically specify.

In our final experiment, we also explore how robustness can be further improved by training on a distribution (ensemble) of training environments, where each environment differs in terms of the physical properties of the manipulated object (its size and mass). By training policies to succeed on the entire ensemble, we can acquire policies that are explicitly trained for robustness, similar in spirit to previously proposed model ensemble methods~\cite{Mordatch15a,Rajeswaran16}.
Interestingly, we observe that for difficult control problems like high dimensional dexterous manipulation, RL from scratch with shaped rewards is unable to learn a robust policy for a diverse ensemble of environments in a time frame comparable to the time it takes to master a single instance of the task. On the other hand, DAPG is still able to succeed in this setting and generates even more robust policies, as shown in Figure~\ref{fig:robustness}.

\section{Conclusion}

In this work, we developed a set of manipulation tasks representative of the types of tasks we encounter in everyday life. The tasks involve the control of a 24-DoF five-fingered hand. We also propose a method, DAPG, for incorporating demonstrations into policy gradient methods. Our empirical results compare model-free RL from scratch using two state-of-the-art DRL methods, DDPG and NPG, as well their demonstration-based counterparts, DDPGfD and our DAPG algorithm. We find that NPG is able to solve these tasks, but only after significant manual reward shaping. Furthermore, the policies learned under these shaped rewards are not robust, and produce idiosyncratic and unnatural motions. After incorporating human demonstrations, we find that our DAPG algorithm acquires policies that not only exhibit more human-like motion, but are also substantially more robust.
Furthermore, we find that DAPG can be up to 30x more sample efficient than RL from scratch with shaped rewards. DAPG is able to train policies for the tasks we considered in under 5 hours, which is likely practical to run on real systems. Although success remains to be demonstrated on real hardware, given the complexity of the tasks in our evaluation and the sample-efficiency of our DAPG method, we believe that our work provides a significant step toward practical real-world learning of complex dexterous manipulation. In future work, we hope to learn policies on real hardware systems, further reduce sample complexity by using novelty based exploration methods, and learn policies from only raw visual inputs and tactile sensing. 

\section*{Acknowledgements}
The authors would like to thank Ilya Sutskever, Wojciech Zaremba, Igor Mordatch, Pieter Abbeel, Ankur Handa, Oleg Klimov, Sham Kakade, Ashvin Nair, and Kendall Lowrey for valuable comments. Part of this work was done when AR interned at OpenAI. AR would like to thank OpenAI for providing a highly encouraging research environment. AG was supported by an NSF fellowship. ET acknowledges funding from the NSF.

\bibliography{references}
\bibliographystyle{abbrv}

\end{document}